# DTCT
## DETECT THEN ACT

**TECHNICAL REPORT 4**

# A FEAST FOR TROLLS
# ENGAGEMENT ANALYSIS OF COUNTERNARRATIVES AGAINST ONLINE TOXICITY

▼▼▼

AUTHORS:
TOM DE SMEDT, PIERRE VOUÉ, SYLVIA JAKI,
EMILY DUFFY, LYDIA EL-KHOURI

This report provides an engagement analysis of counternarratives against online toxicity. Between February 2020 and July 2021, we observed over 15 million toxic messages on social media identified by our fine-grained, multilingual detection AI. Over 1,000 dashboard users responded to toxic messages with combinations of visual memes, text, or AI-generated text, or they reported content. This leads to new, real-life insights on self-regulatory approaches for the mitigation of online hate.

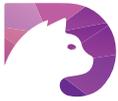

Engagement analysis of
counternarratives against online toxicity

Technical Report 4

# INTRODUCTION

In 2020, more than 10 million new pictures appeared on Facebook and more than 20 million new messages on Twitter every hour. In a study on Dutch hate speech,[1] Textgain discovered at least 500 extremely toxic messages in a random sample of 100,000 tweets (0.5% or 1/200). This potentially means that over a 100K new toxic posts appear on Twitter every hour, and a team of one thousand moderators would need to review 5 posts per minute each, for 8 hours per day, with no lunch breaks or days off. In short, sifting through potentially harmful messages by hand is difficult and onerous. The **Detect Then Act** project (DTCT) funded by the DG Justice of the European Commission uses AI to automatically collect toxic messages in an online dashboard.

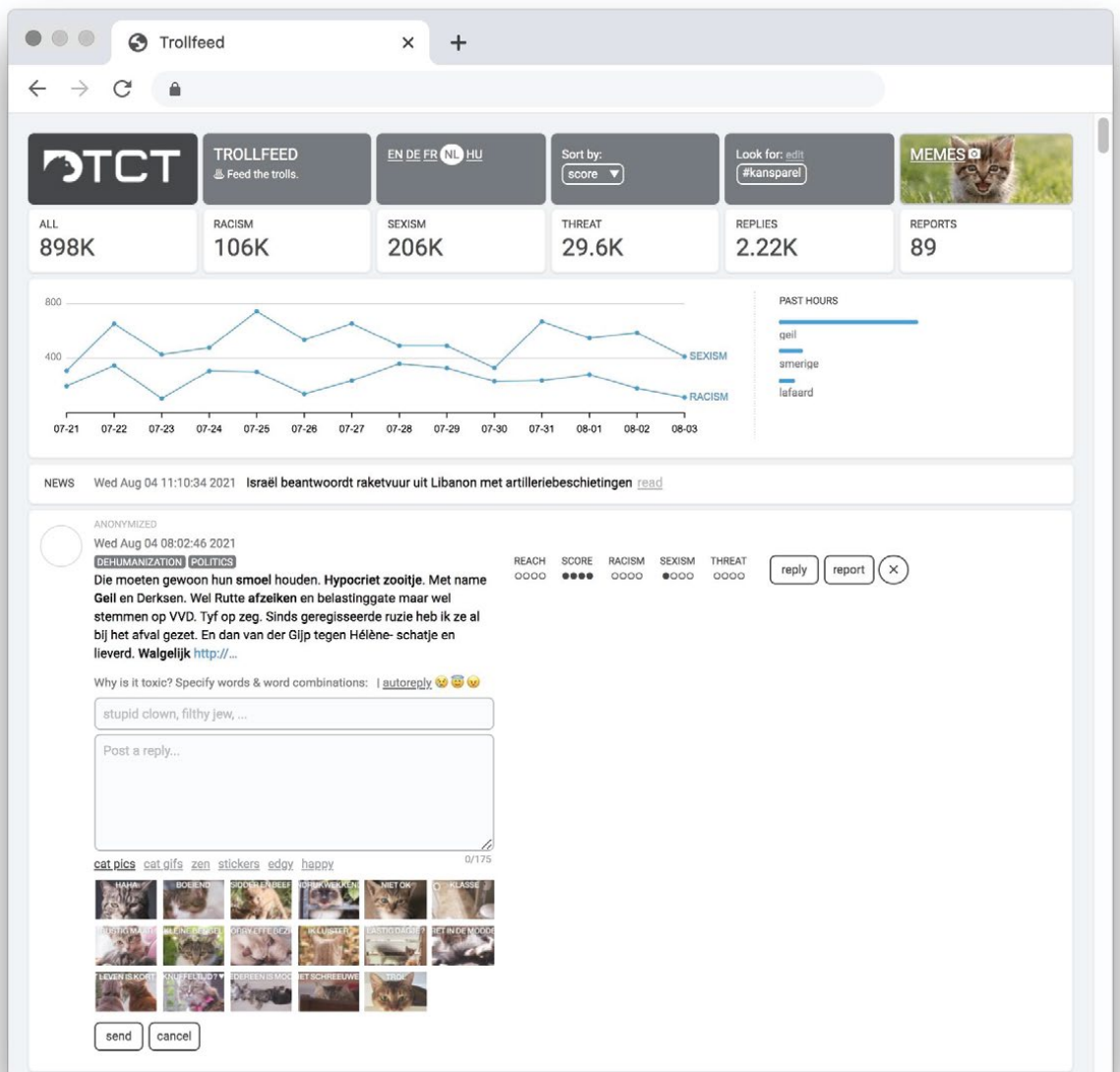

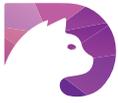
Engagement analysis of
counternarratives against online toxicity
―――
Technical Report 4## 1.1 TOXICITY

Pejorative language use on online social media, and its real-life impact on society, is an active field of study in various academic disciplines, from linguistics to sociology and political science. Most often, the language involved is referred to as *hate speech*, or sometimes also as *abusive language, harmful language, offensive language, cyberbullying or cyberhate*. These denominations can be complicated, since they are not very well defined, and imply that there always is a target being hated or abused. Social media users post unusual things in a wide variety of circumstances. They may be upset, they may be having fun, and/or they may not even be aware of the impact of their words on others.

We focus on **toxicity**, where *toxic* means language use on a scale from unnecessary to unpleasant to unacceptable, with a spectrum of manifestations such as profanity, ridicule, verbal aggression and targeted hate speech.[2/3] The scale can be represented as a number, and the spectrum as an open-ended combination of fine-grained labels (e.g., RACISM, SEXISM, THREAT).

## 1.2 TOXICITY DETECTION

Essentially, our Explainable AI acts as a virtual assistant, imitating what humans would do. Under the hood, it knows about thousands of offensive words and word combinations (*fat bitches, filthy pigs, faggot retards, hordes of immigrants* etc.), and continually searches social media for new messages that contain such expressions, without lunch breaks or days off. Human experts in linguistics and media communication trained the AI about what words are worse than others, which words express racism, sexism, aggression, which are dehumanizing, and so on.[4/5]

Harnessing this knowledge, the AI will then attempt to assist dashboard users by presenting what it believes are the worst cases. The identity of the author of a toxic message will not be revealed in the dashboard, since EU privacy regulations (GDPR) rightly forbid tracking and collecting information about individuals. This works in both directions: users of the dashboard don't know the authors of toxic messages, and those authors won't know the users that post a response using the dashboard.

Dashboard users can choose which messages to act on, by writing a counternarrative **response** or **reporting** them depending on the laws applicable to different countries. Over the course of the project monitoring (February 2020 till July 2021) over 1,000 users supervised by expert trainers were involved. These trainers were in turn supervised by first-line practitioners and academic experts from linguistics, political science, communication science, and arts. We monitored the counternarratives that were posted and their engagement rate, resulting in real-life insights.

[2] https://dl.acm.org/doi/10.1145/3308560.3317083
[3] https://dl.acm.org/doi/pdf/10.1145/2818052.2869107
[4] https://www.textgain.com/wp-content/uploads/2020/06/TGTR3-pow.pdf
[5] https://www.textgain.com/wp-content/uploads/2020/03/TGTR1-48ch.pdf3

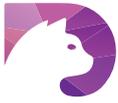

# 2.1 DISTRIBUTION OF TOXICITY BY LANGUAGE

Table 1 provides a breakdown of toxicity by language over a 1.5 year period. For every language, about every two minutes the AI will search social media for messages that contain toxic keywords and that were posted in the past two days. It will rank these by level of toxicity, keep the 2,000 most toxic ones in memory, and present a selection to dashboard users. By consequence, this means that there are many more toxic messages seen by the AI than those that the users will ever see, as the 'pool' continually fills up with newer, more toxic messages replacing older, less toxic messages.

| LANGUAGE | SEEN | RACISM | SEXISM | THREAT | REPLIES | REPORTS | REMOVALS |
|---|---|---|---|---|---|---|---|
| EN | 6M | 1.5M | 750K | 500K | 1K | 750 | 45% |
| DE | 2M | 250M | 300K | 75K | 750 | 125 | 50% |
| FR | 6M | 1.5M | 1.5M | - | 25 | 125 | 25% |
| NL | 1M | 100K | 200K | 25K | 2K | 100 | 30% |
| HU | 125K | 25K | 25K | - | 50 | 250 | 10% |

**Table 1.** Toxic messages by language

For example, during the 1.5 year period the AI identified about 6M toxic messages written in English, of which it labeled about 1/4 as racist, about 1/8 as sexist and 1/12 as threatening. Dashboard users responded to about 1,000 of these and reported about 500, of which about half were effectively removed by the social media platform. The most removals occurred in German content, likely because the country's NetzDG law obliges social media companies to act on reported content within 24 hours. The most replies were posted to Dutch toxic messages, while at the same time Dutch toxic messages were less reported and removed.

In the following sections, we will mainly focus on the English, German and Dutch messages. To offer a general idea for English, messages that dashboard users tend to respond to frequently contain toxic keywords such as *bitch* (500x), *fucking <noun>* (250x), *scum* (150x), *ugly* (150x), and *stupid* (125x), in other words content that the AI typically marks as PROFANITY and RIDICULE. On the other hand, messages that dashboard users tend to report frequently contain toxic keywords such as *nigger* (350x), *kike*, *faggot*, *nazi*, etc., which often indicate discriminatory or RACIST content.

## 2.2 DISTRIBUTION OF TOXICITY BY LABEL

Figure 1 provides a breakdown of toxicity by label. For example, for messages in English, German and Dutch that are identified as toxic, about 1/4 is labeled as RACISM (19.5 + 3.5 + 1.5 ≈ 25%). About 1/4 of what the AI identifies as toxic, and subsequently presents to dashboard users, is PROFANITY, in other words messages that contain expressions such as *shit* and *fuck* without necessarily targeting an out-group. About 1/4 is labeled as RIDICULE, in other words messages that contain expressions such as *idiot* and *clown* without necessarily discriminating on the basis of ethnicity or gender.

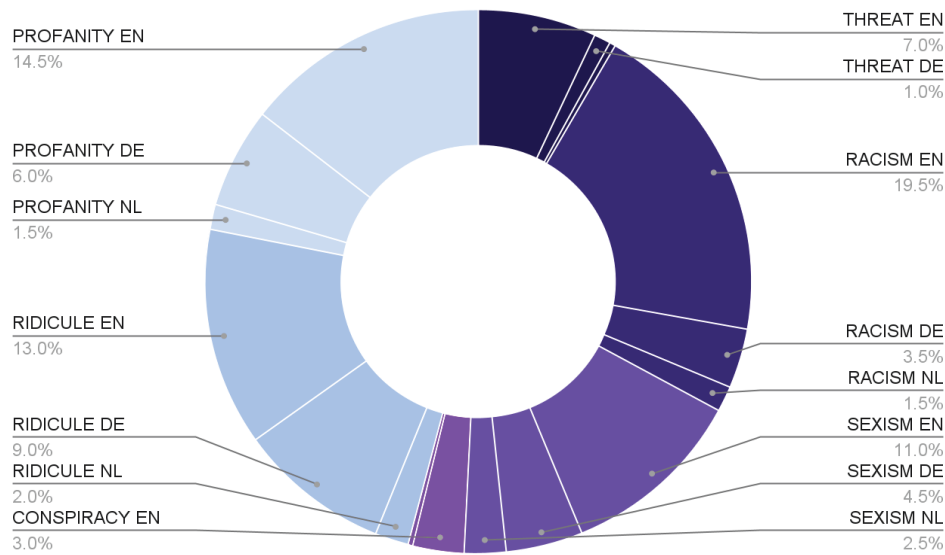

**Figure 1.** Toxic messages by label

Toxic messages may have multiple labels, like RACISM + THREAT. Messages labeled as threats contain expressions such as *fight*, *shoot* and *kill*. The combination of racist slurs and verbal aggression usually results in very high toxicity scores, and hence such messages will appear at the top of the list in the dashboard, where they can be seen by many users. These messages often constitute content that breaks the social media platforms' Terms of Service (ToS), or violates local hate speech legislation.

In effect, about 10% of messages that are reported by users (instead of being replied to) are labeled as RACISM + THREAT. About 50% of reported messages were labeled as RACISM, and 25% as THREAT. In short, after going through the training programme,[6] users correctly focus on reporting content that should be removed by the social media platforms, and use self-regulatory approaches to raise awareness by responding with counternarratives to other content. A later addition was the detection of CONSPIRACY messages, which was insightful in the course of the Donald Trump presidency and the COVID-19 pandemic (see also section 2.2).

[6] https://dtct.eu/resources

# 2.3 DISTRIBUTION OF TOXICITY BY LANGUAGE OVER TIME

Figure 2 provides a breakdown of toxicity by language over time. For example, messages in English that are identified as toxic reach a peak around February 13. These peaks are not coincidental but correlate to media coverage about societal tensions. Some interesting examples are given below.

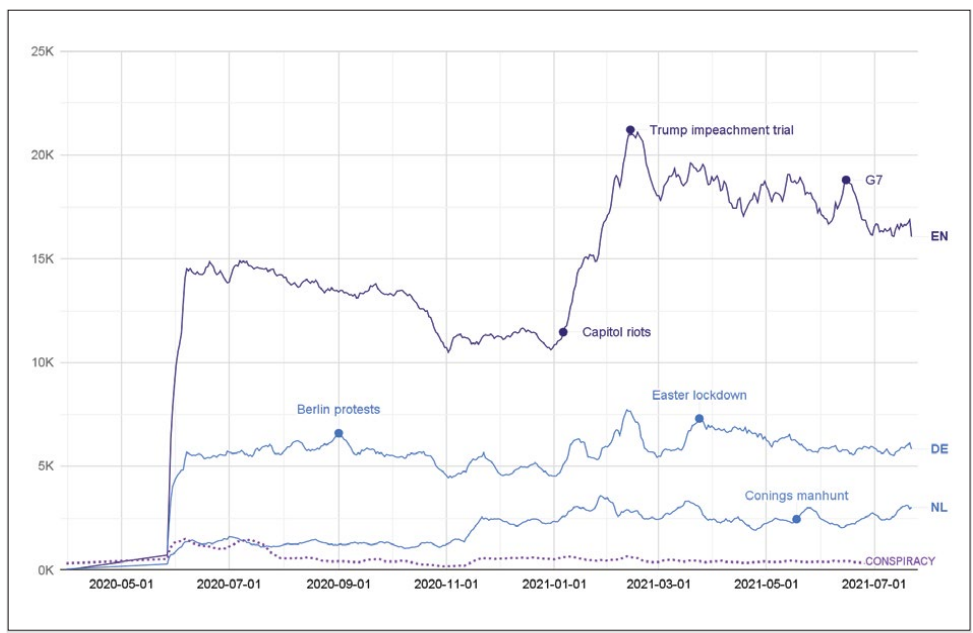

**Figure 2**. Toxic messages by language over time

Essentially, the timeline shows three development stages of the detection AI. Recording its activity begins around June 2020 (1), followed by a period with few peaks until November 2011. The reason for the seemingly uneventful curves in this stage is that the detection AI is scanning social media for toxic keywords (e.g., *idiot*, *clown*) with no particular goal. By consequence, a lot of noise is picked up about any kind of discussion topic large and small, even though notable events such as the storming of the Reichstag by far-right protesters in Germany do surface. From November 2020 to January 2021 (2), the AI underwent a number of updates in collaboration with the dashboard users. In particular, it started to digest news articles, to pick up societal tensions.

From then on (3), the AI searches social media for combinations of trending topics and toxic keywords (e.g., *Trump + idiot*).

The effect is noticeable: the AI becomes more sensitive to peaks, which represent social media users responding to media coverage of world events that upset them in some way. This resulted in a much more efficient dashboard. Instead of displaying random toxic messages, for example a friend group discussing their computer game with PROFANITY and RIDICULE, it now displays social media fallout about real-time events, and it becomes a tool for studying and self-regulating public outrage.

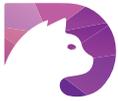

Engagement analysis of
counternarratives against online toxicity
———
Technical Report 4

**Short discussion of a few peaks:**

**Berlin protests**: on August 29, 2020, hundreds of allegedly far-right protesters attempt to 'storm' the German Reichstag parliament building.[7] Some of the protesters carry QAnon flags, others include conspiracists and anti-vaxxers.[8] This event correlates with a peak of toxic messages in German, in particular with 2x as many CONSPIRACY messages as usual.

**Capitol riots**: on January, 6, 2021, thousands of Trump supporters breach the US Capitol building in support of the president's claim that the election was 'stolen', carrying QAnon flags.[9] The backlash on social media, with intensely polarized discussion between left-wing and right-wing voters, correlates with a rising peak of toxic messages in English, but also in German and Dutch. Of note is the CONSPIRACY curve at the bottom (dotted line). It is highest in summer 2020, essentially when QAnon and coronavirus conspiracy theories are running rampant, and it is at its lowest after March 2023, essentially when Donald Trump and thousands of other QAnon-related accounts are removed from social media platforms,[10] while government vaccination campaigns begin to pay off.

**Trump impeachment trial**: on February 9, 2021, following the Capitol riots, Trump is impeached for the second time and acquitted on February 13,[11] correlating with the highest peak of toxic messages in English. In the following months, we can very slowly see the curve to be dropping again.[12]

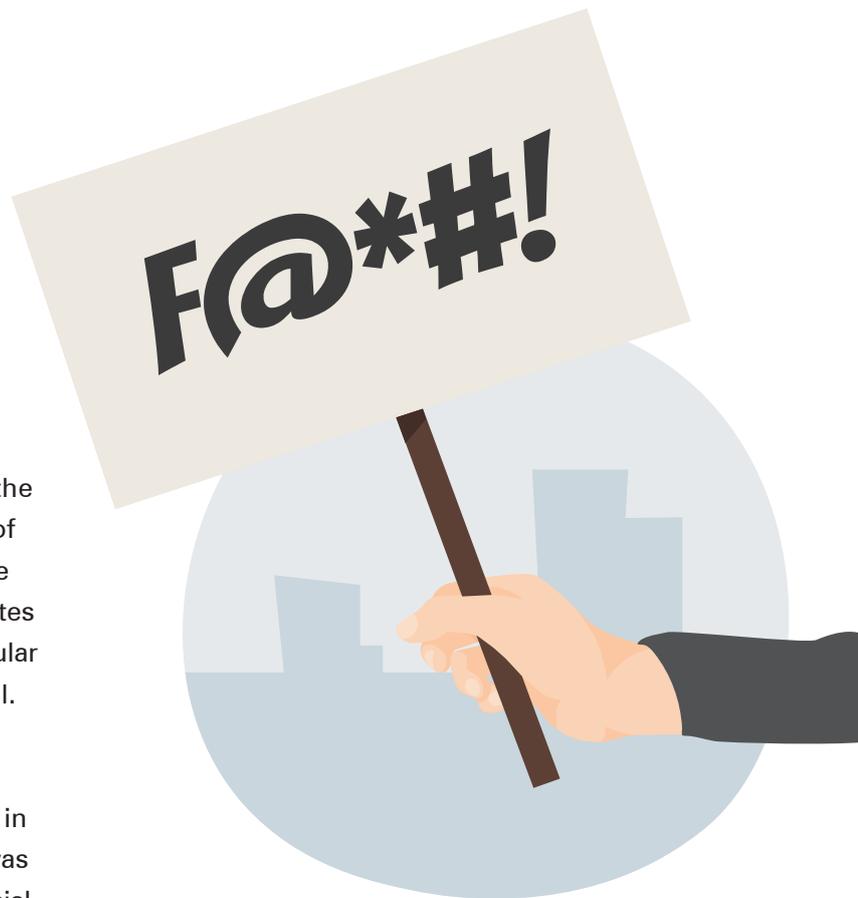

**Easter lockdown:** on March 23, 2021, German chancellor Angela Merkel announces that coronavirus lockdown measures will be extended over Easter for another month. But after a day of criticism and confusion, the plan is cancelled again. This correlates with a spike of toxic messages in German.

**Conings manhunt**: on May 18, 2021, a far-right Belgian soldier goes missing, after taking four rocket launchers from the military barracks and leaving violent threats directed at virologists. This marks the start of a peak of toxic messages in Dutch, as far-right sympathizers of the so-called 'Rambo 2.0' or 'modern Robin Hood' shout out their support. The peak drops again by the end of May, after social media platforms remove a number of controversial support groups and accounts.[13]

---

[7] https://www.bbc.com/news/world-europe-germany-coronavirus-anger-after-attempt-to-storm-parliament
[8] https://www.researchgate.net/publication/the-qanon-superconspiracy
[9] https://www.dw.com/en/us-capitol-storming-what-you-need-to-know
[10] https://theconversation.com/facebook-youtube-moves-against-qanon
[11] https://apnews.com/article/donald-trump-capitol-siege-riots-trials-impeachments
[12] https://www.dw.com/en/covid-angela-merkel-backtracks-on-easter-lockdown-after-uproar
[13] https://www.politico.eu/article/facebook-takes-down-support-group-for-belgian-extremist-soldier

# 3.1 ENGAGEMENT RATE OF COUNTERMEMES

Figure 3 provides an overview of 15 memes for which engagement was tracked. The dashboard has about 50 memes and 50 gifs made by art students in Belgium, the Netherlands, Poland and Finland, from which users can pick one to respond to toxic messages. The subset was chosen for its consistent visual style, to eliminate the variable of aesthetic preference from the measurement. The subset has three communication styles: **RIDICULING** memes that combine a positive text (*fascinating!*) with a negative image (*cat is bored and sleeping*), **REPROACHING** memes that combine a negative text (*troll!*) with a negative image (*cat stares accusingly*), and **RECONCILING** memes that combine a positive text (*we're all purfect*) with a positive image (*different-looking cats chilling*):

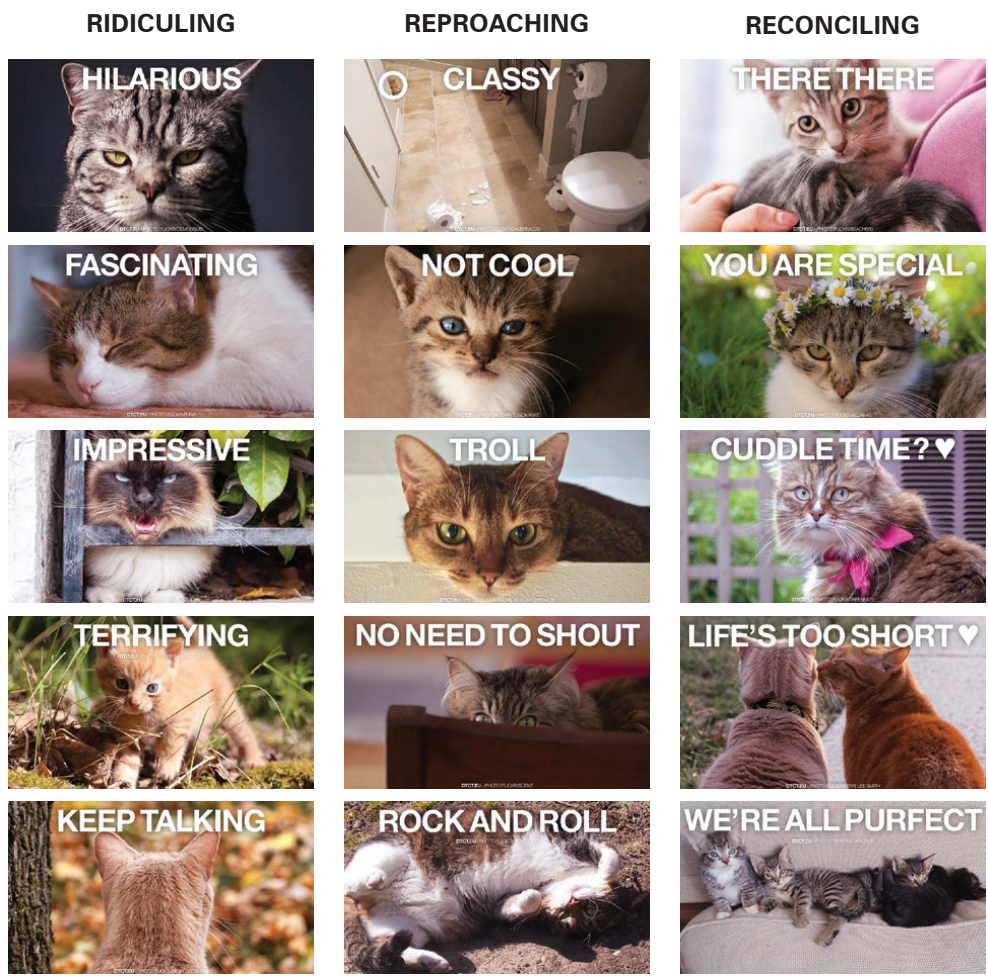

**Fig 3a.** RIDICULING memes have a positive text and negative image.

**Fig 3b.** REPROACHING memes have both a negative text and image.

**Fig 3c.** RECONCILING memes have both a positive text and image.

Each of these cat memes has a translation in EN, DE, FR, NL, HU, to ensure our intent comes across correctly when responding to messages in these languages. The 'cat theme' was chosen because it doesn't depict actual people or events, i.e., it is a metaphor, which should further reduce the chance of unintentional offense when posting the memes. Then, we can measure if there is any difference between the three communication styles. About 3K cat memes were posted in the 1.5 year period.

Dashboard users picked RIDICULING memes (45%) more often than REPROACHING memes (30%) and RECONCILING memes (35%). In particular, the *not cool* meme is the most popular. It is used in 1/3 of all responses that include a cat meme, and it is twice as popular as the *there there* meme, followed by the *cuddle time, no need to shout* and *classy memes*. However, these are not necessarily the most effective memes. In terms of user engagement, we tracked the number of likes each response gets on social media, as an indicator of bystanders supporting the counternarratives.

The engagement rate of the most popular *not cool* meme is only 3%. This means that less than 1/30 of these get liked. Other memes with a low engagement rate are *troll* (2%), *no need to shout* (4%) and *there there* (4%) – the latter of which arguably belongs in the RIDICULE category instead of in the RECONCILING category. The most effective is the *life's short* meme with an engagement rate of nearly 10%, followed by the *purfect* and *cuddle* memes (8%), and the *classy* and *fascinating* memes (7%).

In general, RIDICULING memes have an engagement rate of about 5%, REPROACHING memes 4%, and RECONCILING memes 8%. Intuitively, this makes sense. It makes sense for dashboard users to prefer more confronting memes, which may offer a feeling of comfort or accomplishment that something real was done against an online troll. It also makes sense that the art students, when left to their own ideas, prefer to create confronting memes, since it is more exciting. However, the aim of the study is not to confront the trolls, if only because there are real people behind virtual accounts, but rather to engage more bystanders to help mitigate online polarization. It makes sense that bystanders would express support in toxic discussions that seem to have a solution, featuring **reconciling memes**, and would shy away from toxic discussions that have no end in sight, featuring retaliating memes.[14]

---

[14] https://en.wikipedia.org/wiki/Tit_for_tat

## 3.2 ENGAGEMENT RATE OF COUNTERNARRATIVES

The dashboard uses a double-blind approach, meaning that the authors of toxic messages are anonymized,[15] and responses to the messages are posted from the anonymous @trollfeeders account to protect the identity of the users of the dashboard. This has important advantages but also disadvantages.

For one, as shown in Figure 4a, users will have to base their responses on the actual content of toxic messages, without knowing the author. Hence, all authors are treated equally, even if it is Geert Wilders, the leader of the Dutch right- wing nationalist Party for Freedom (PVV). The biggest disadvantage in general is that people tend to see anonymous accounts as a nuisance and as less important. A response that is often posted in reply to the counternarratives is "and who are you?" or "yes but why are you here?". However, these disadvantages do not weigh up to protecting the identity of the users of the dashboard against doxing on 4chan.[16]

The second advantage is that the @trollfeeders account is announced and known to the social media platforms, which establishes a relation of trust that increases the chance of reported messages being reviewed more closely, while also getting access to platform's API to monitor interactions on the account automatically (e.g., analysing engagement).

Figure 4b shows one example of engagement, where one person chooses to respond to the *there there* meme with a like and a "that is an adorable fucking cat", effectively defusing the toxic discussion.

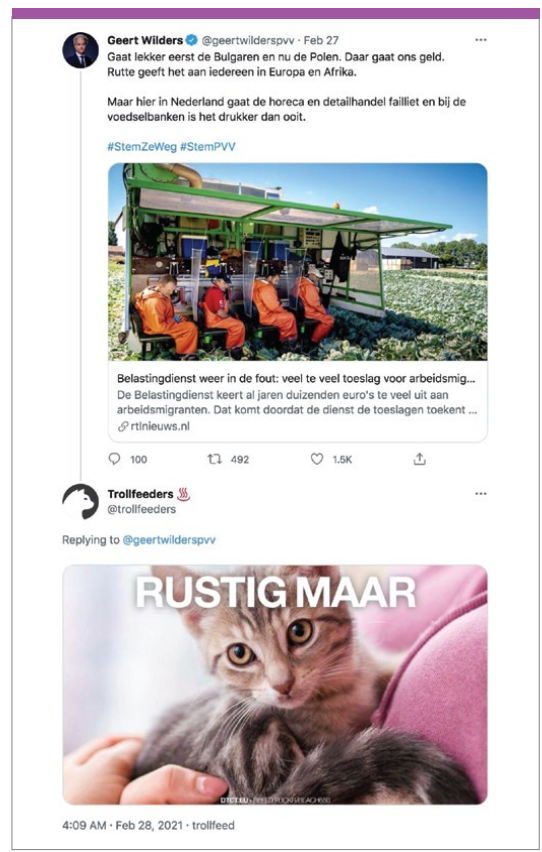

**Figure 4a**. Example counternarrative.

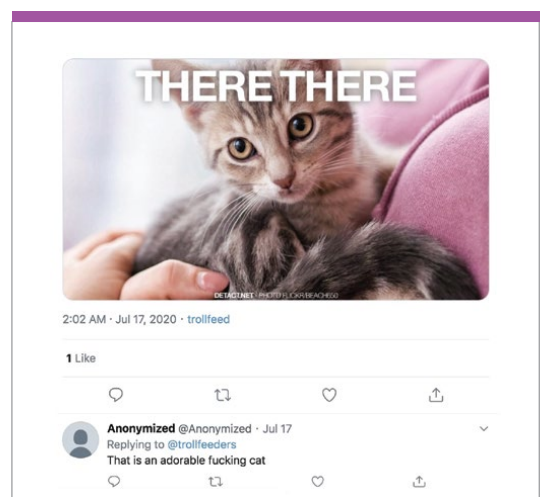

**Figure 4b**. Example counternarrative.

---

[15] https://gdpr.eu
[16] https://www.theguardian.com/2018/jun/14/doxxing

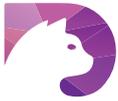

Engagement analysis of counternarratives against online toxicity

Technical Report 4

Dashboard users can respond to messages with a meme, a text, a text co-written by the AI, or any combination of those. The AI can currently compose satisfactory responses in English and Dutch. In brief, it uses a Context-free Grammar of replacement rules.[17] For example, in "please be @polite", the @polite placeholder could be replaced with *nice*, *polite*, *respectful*, etc.

A simple sentence with three placeholders that each have ten possible replacements leads to 1,000 variations of the same narrative. Along with meta-information (e.g., is this a response to RACISM?) AI responses are quite convincing. Figure 4c shows one interesting example of an author of a toxic message engaging with the AI, and reflecting on what they wrote. Figure 4d is an unsuccessful example of an AI response.

Dashboard users posted about 4K responses to toxic messages, of which 3.5K (85%) contain a meme, 2.5K (60%) contain a text, and 1K (25%) contain a text written by the AI. However, the AI assistant was only introduced in the last six months, and is quickly becoming more popular. The reason is that it is easier to scroll through possible responses generated by the AI than it is to mentally engage with a toxic message.

For English and Dutch, counternarratives that only contain a meme are almost never liked, those that only contain an AI-written text are liked 0.5% of the time, while those that only contain a human-written text are liked 2% of the time. Counternarratives with a meme and a human-written text are the most effective, with a 5% engagement rate (1/20). However, a combination of a meme with an AI-written text is not far behind, with a 3% engagement rate.

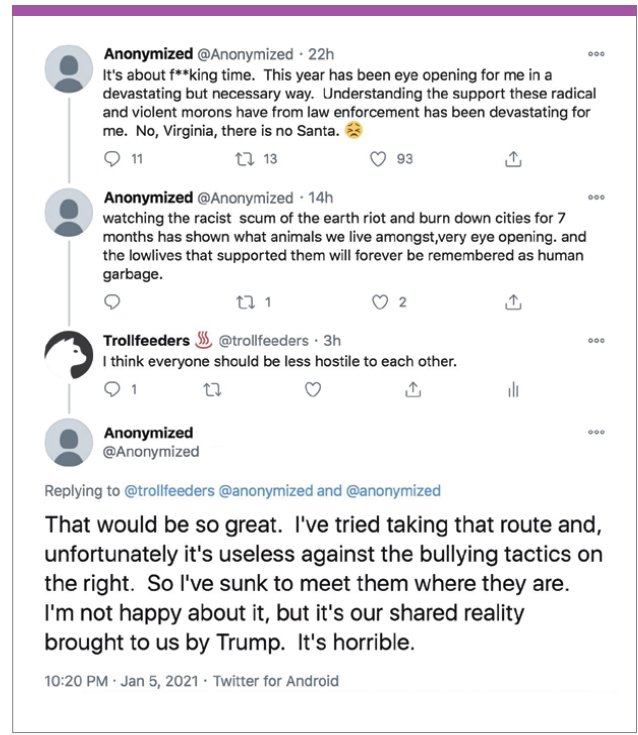

**Figure 4c**. Example AI counternarrative.

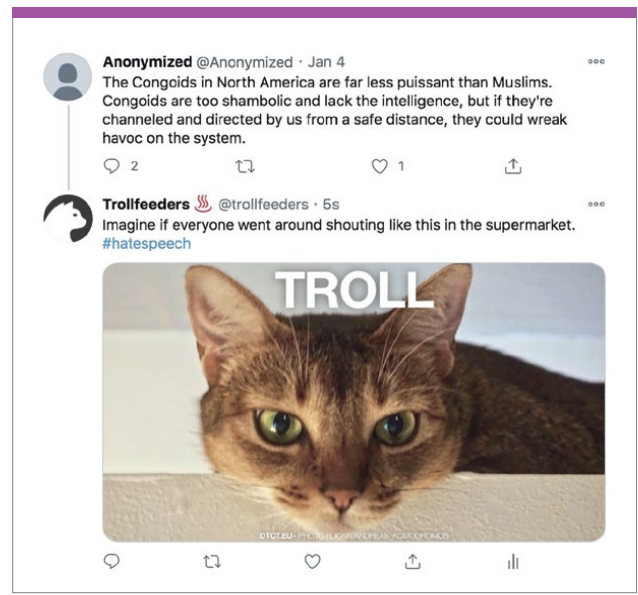

**Figure 4d**. Example AI counternarrative.

---

[17] https://en.wikipedia.org/wiki/Context-free_grammar

## 3.3 REPORTED MESSAGES VS REMOVED MESSAGES

Table 1, section 2.1 shows the number of toxic messages reported by dashboard users, and the percentage of those messages that were removed by the social media platform. About 1.25K messages were reported in total, and about 500 (40%) were also removed. The large part of messages were reported through the dashboard, and some manually by the project managers. These include 1 jihadist account that issued death threats on the night of the Vienna shootings on November 2, 2020,[18] 5 neo-Nazi accounts with holocaust denial, and 2 death threats against popular virologists. These were also reported to law enforcement. We also manually reported about 10 violent racist messages targeting Bukayo Saka during the Euro 2020 football finale.[19]

In the earlier months of the project, about 50 messages per month were reported, rising up to about 150 (3x) messages per month in the later months. By the end of the project, there were about 1,000 dashboard users – much more than the 100 which were along from the very beginning. This reflects in the reported vs removed ratio. In the earlier months, nearly 60% of reported messages were removed, but in the later months, only 30% of reported messages were indeed removed. The reason is that the first 100 users received much closer, hands-on, in-person training, while the larger group only had the project training manuals and decision trees to work with (Figure 6).

This highlights the importance of the need for more media literacy education.

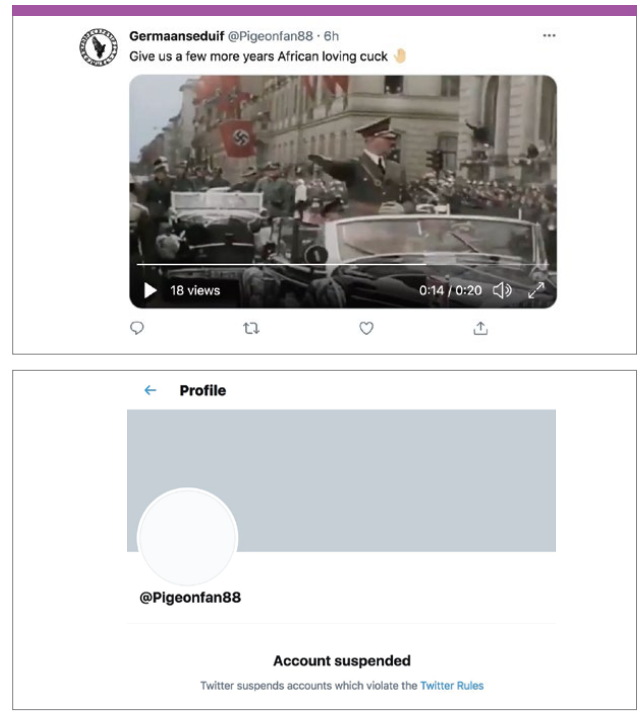

**Figure 5**. Example reported content.

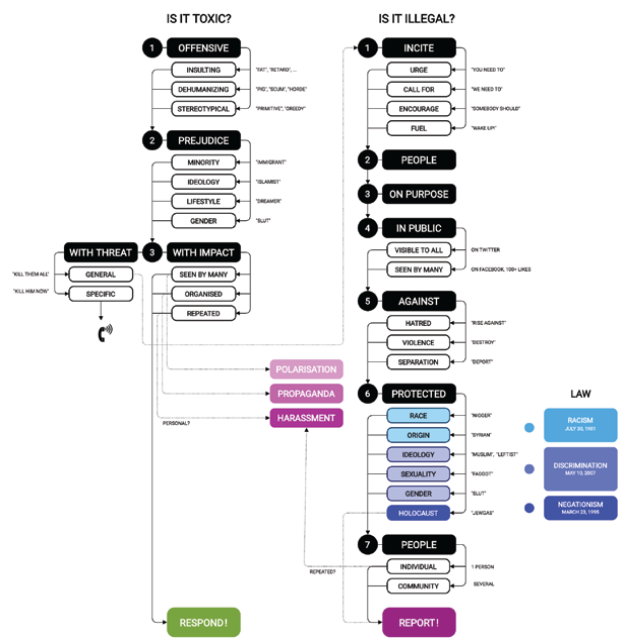

**Figure 6**. Decision tree for NL reporting.

---

[18] https://www.bbc.com/news/world-europe-54798508
[19] https://www.bbc.com/news/uk-57848761

# 4.1 PERSPECTIVES ON EXPLAINABLE AI

Typically, more complex Machine Learning algorithms (ML) produce models that are more accurate, but this may come at a price. More complex algorithms are also more difficult to explain, and the predictions made by resulting models difficult to interpret. Hence, such models are sometimes called 'black boxes'. When applied in real-life in high-risk applications, e.g., predicting cancer or labeling someone as a potential jihadist, it can lead to ethical and legal challenges. If we do not know how a model functions, we cannot know if it is somehow prejudiced, presenting a danger to human rights.

Several studies have already warned against using black box models for high-risk decision making,[20] and future EU regulations may clamp down on such applications.[21] In DTCT, we believe that AI must always be supervised by humans. Our AI search algorithm does not track or store private information in compliance with the GDPR, our AI ranking algorithm highlights potentially toxic messages but defers any action (respond / report / ignore) to humans, and our AI language generator diversifies on counternarratives written by dashboard users that had a high engagement rate – it does not invent its own narratives independently. Dashboard users control it directly in a simple interface (Figure 7).

Exciting work is happening in the domain of improving the effectiveness of counternarratives. New datasets are emerging,[22] as well as new strategies to collect such datasets automatically.[23] The next step would be to create AI systems that respond to toxic messages independently, leveraging large language models like GPT-2. However exciting, this presents a new risk in the form of perpetuating historical prejudice. To see what we mean, simply enter "the jew women will" in an online GPT-2.[24]

**Figure 7.**

---

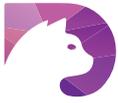

## 4.2 PERSPECTIVES ON SELF-REGULATION

One of the most requested features by dashboard users is the ability to observe the **context** in which a toxic message was posted – basically the message it was replying to. To illustrate this, one reply with 14 likes to one of our counternarratives read: "@trollfeeders Bist du noch ganz dicht? Was für rassismus? 😄 Mein onkel ist Türke. 👨 Nicht alles nachplappern!" ("Trollfeeders, are you dim? What racism? My uncle is Turkish. Stop parroting!"). We missed the context in which the discussion was taking place. However, the solution is not that simple.

On average, our AI search algorithm does one query every 20 seconds (e.g., any new messages that contain the word *idiot*), resulting in hundreds of identified toxic messages by our AI ranking algorithm. If we would have to retrieve the context for every one of these, we would need to do thousands of extra search queries every minute. There is currently no social media platform API that allows this kind of volume. Perhaps the EU's intention to create a framework for 'trusted flaggers' in the Digital Services Act will offer new perspectives, fostering tech companies and humanitarian initiatives to work more closely, with extended API access.

### Qualitative insights for English content

We conducted a qualitative analysis on 1,000 English toxic messages that dashboard users responded to or reported. The most common types of toxicity that trigger users to act are racism and misogyny, followed by conspiracy theories (which were abundant during coronavirus lockdown measures), and to a lesser extent Islamophobia and fat-shaming. In about half of the cases, trolls used container metaphors (e.g., *they keep streaming in*), followed by animal metaphors (e.g., *they are like vermin*, 20%) – in particular comparisons to *sheep* (5%) and *parasites* (5%) – and free speech metaphors (e.g., *the people are being muzzled*).[25] Roughly 1/3 of responses posted by dashboard users employ some form of sarcasm or ridicule, which our engagement analysis suggests is not necessarily that effective. In effect, prior research shows that 'silent' readers find sarcasm more often negative than funny.[26] When people see two groups in conflict, they tend to isolate both of them. One way forward is to focus more on shared commonalities instead of differences, which is discussed in section 3.

---

[25] https://minerva.usc.es/xmlui/bitstream/handle/10347/20167
[26] https://kids.frontiersin.org/articles/10.3389/frym.2018.00056

## ACKNOWLEDGEMENTS

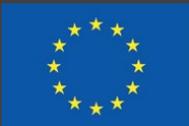
This project is co-funded by the Rights, Equality and Citizenship Programme of the European Union (2019-2021). The content of this report represents the views of the author only and is his/her sole responsibility. The European Commission does not accept any responsibility for use that may be made of the information it contains.

DTCT
DETECT THEN ACT